\documentclass{article} 
\usepackage{iclr2016_conference,times}
\usepackage{hyperref}
\usepackage{url}
\usepackage{tabularx}

\usepackage{amsmath, amssymb, amsfonts}
\newcommand{\bx}{\mathbf{x}}

\newcommand{\bz}{\mathbf{z}}

\newcommand{\by}{\mathbf{y}}

\newcommand{\cLD}{\mathcal{L}_D}
\newcommand{\cLG}{\mathcal{L}_G}

\definecolor{dgreen}{rgb}{0,.7,0}
\definecolor{dyellow}{rgb}{.7,.7,0}
\definecolor{dred}{rgb}{.7,0,0}
\definecolor{dblue}{rgb}{0,0,0.7}

\newcommand{\TODO}[2][JTS]{}

\usepackage{graphicx} 

\usepackage{multirow}

\title{Unsupervised and Semi-supervised Learning with Categorical Generative Adversarial Networks}

\author{Jost Tobias Springenberg \\
University of Freiburg \\
79110 Freiburg, Germany\\
\texttt{springj@cs.uni-freiburg.de} \\
}

\iclrfinalcopy 

\begin{document}

\maketitle

\begin{abstract}
  In this paper we present a method for learning a discriminative
  classifier from unlabeled or partially labeled data.  Our approach
  is based on an objective function that trades-off mutual
  information between observed examples and their predicted
  categorical class distribution, against robustness of the classifier
  to an adversarial generative model. The resulting algorithm can
  either be interpreted as a natural generalization of the generative
  adversarial networks (GAN) framework or as an extension of the
  regularized information maximization (RIM) framework to robust
  classification against an optimal adversary.  We empirically
  evaluate our method -- which we dub categorical generative
  adversarial networks (or CatGAN) -- on synthetic data as well as on
  challenging image classification tasks, demonstrating the robustness
  of the learned classifiers.  We further qualitatively assess the
  fidelity of samples generated by the adversarial generator that is
  learned alongside the discriminative classifier, and identify links
  between the CatGAN objective and discriminative clustering
  algorithms (such as RIM).
\end{abstract}

\section{Introduction}
Learning non-linear classifiers from unlabeled or only partially
labeled data is a long standing problem in machine learning. The
premise behind learning from unlabeled data is that the structure
present in the training examples contains information that can be used
to infer the unknown labels. That is, in unsupervised learning we
assume that the input distribution $p(\bx)$ contains information about
$p(y \mid \bx)$ -- where $y \in \lbrace 1, \dots, K \rbrace$ denotes
the unknown label.  By utilizing both labeled and unlabeled examples
from the data distribution one hopes to learn a representation that
captures this shared structure. Such a representation might,
subsequently, help classifiers trained using only a few labeled
examples to generalize to parts of the data distribution that it would
otherwise have no information about. Additionally, unsupervised
categorization of data is an often sought-after tool for discovering
groups in datasets with unknown class structure.

This task has traditionally been formalized as a cluster assignment
problem, for which a large number of well studied algorithms can be
employed. These can be separated into two types: (1)
\emph{generative clustering} methods such as Gaussian mixture models,
k-means, and density estimation algorithms, which directly try to model
the data distribution $p(\bx)$ (or its geometric properties); (2)
\emph{discriminative clustering} methods such as maximum margin
clustering (MMC) \citep{MMC_2004} or regularized information
maximization (RIM) \citep{Krause_2010}, which aim to directly group
the unlabeled data into well separated categories through some
classification mechanism without explicitly modeling $p(\bx)$. While
the latter methods more directly correspond to our goal of learning
class separations (rather than class exemplars or centroids), they can
easily overfit to spurious correlations in the data; especially when
combined with powerful non-linear classifiers such as neural networks.

More recently, the neural networks community has explored a large
variety of methods for unsupervised and semi-supervised learning
tasks. These methods typically involve either training a generative
model -- parameterized, for example, by deep Boltzmann machines
(e.g. \citet{SalHinton07}, \citet{Goodfellow_NIPS2013}) or by
feed-forward neural networks (e.g. \citet{Bengio_ICML2014},
\citet{Kingma_NIPS2014}) --, or training autoencoder networks
(e.g. \citet{HintonSalakhutdinov2006b},
\citet{VincentPLarochelleH2008}). Because they model the data
distribution explicitly through reconstruction of input examples, all
of these models are related to generative clustering methods, and are
typically only used for pre-training a classification network.  One
problem with such reconstruction-based learning methods is that, by
construction, they try to learn representations which preserve all
information present in the input examples. This goal of perfect
reconstruction is often directly opposed to the goal of learning a
classifier which is to model $p(y | \bx)$ and hence to only preserve
information necessary to predict the class label (and become invariant
to unimportant details)

The idea of the categorical generative adversarial networks (CatGAN)
framework that we develop in this paper then is to combine both the
generative and the discriminative perspective. In particular, we learn
discriminative neural network classifiers $D$ that maximize mutual
information between the inputs $\bx$ and the labels $y$ (as predicted
through the conditional distribution $p(y | \bx, D)$) for a number
of $K$ unknown categories. To aid these classifiers in their task of
discovering categories that generalize well to unseen data, we enforce
robustness of the classifier to examples produced by an adversarial
generative model, which tries to trick the classifier into accepting
bogus input examples.

The rest of the paper is organized as follows: Before introducing our
new objective, we briefly review the generative adversarial networks
framework in Section \ref{sect:gan}. We then derive the CatGAN
objective as a direct extension of the GAN framework, followed by
experiments on synthetic data, MNIST~\citep{LeCun_NC1989} and
CIFAR-10~\citep{Krizhevsky2009}.

\TODO{I currently interchangeably use samples and examples, I should
  really change notation to be consistent! }

\TODO{Also check and make uses of category vs. class consistent.}

\TODO{unsupervisedly ?}

\section{Generative adversarial networks}
\label{sect:gan}
Recently, \citet{Goodfellow_NIPS2014} introduced the generative adversarial
networks (GAN) framework. They trained generative models through an
objective function that implements a two-player zero sum game between
a discriminator $D$ -- a function aiming to tell apart real from fake
input data -- and a generator $G$ -- a function that is optimized to generate
input data (from noise) that ``fools'' the discriminator. The ``game''
that the generator and the discriminator play can then be intuitively
described as follows. In each step the generator produces an
example from random noise that has the potential to fool the
discriminator. The discriminator is then presented a few real data
examples, together with the examples produced by the generator, and
its task is to classify them as ``real'' or ``fake''. Afterwards, the
discriminator is rewarded for correct classifications and the generator
for generating examples that did fool the discriminator. Both
models are then updated and the next cycle of the game begins.

This process can be formalized as follows. Let $\mathcal{X} =
\lbrace \bx^1, \dots \bx^N \rbrace$ be a dataset of provided ``real'' inputs
with dimensionality $I$ (i.e. $\bx \in \mathbb{R}^I$). Let
$D$ denote the mentioned discriminative function and $G$ denote the
generator function. That is, $G$ maps random vectors $\bz \in
\mathbb{R}^Z$ to generated inputs $\tilde{\bx} = G(\bz)$ and we assume
$D$ to predict the probability of example $\bx$ being present in the
dataset $\mathcal{X}$: $p(y = 1 
\mid \bx, D) = \frac{1}{1 + e^{-D(\bx)}}$. The GAN objective is then given as
\begin{equation}
\label{eq:gan_loss}
        \min_{G}
  \max_{D} \ \mathbb{E}_{\bx \sim \mathcal{X}} \Big[
                          \log p(y=1 \mid \bx, D)
  \Big] 
        + \ \mathbb{E}_{\bz \sim P(\bz)} \Big[ \log \big ( 1 - 
                          p(y=1 \mid G(\bz),
                          D) \big )
               \Big],
\end{equation}
where $P(\bz)$ is an arbitrary noise distribution which -- without loss of
generality -- we assume to be the uniform distribution $P(z_i) =
\mathcal{U}(0,1)$ for the remainder of this paper. If both the generator
and the discriminator are differentiable functions (such as deep neural
networks) then they can be trained by alternating stochastic
gradient descent (SGD) steps on the objective functions from Equation
\eqref{eq:gan_loss}, effectively implementing the two player game
described above.

\section{Categorical generative adversarial networks (CatGANs)}
Building on the foundations from Section \ref{sect:gan} we will now
derive the categorical generative adversarial networks (CatGAN)
objective for unsupervised and semi-supervised learning. For the
derivation we first restrict ourselves to the unsupervised setting,
which can be obtained by generalizing the GAN framework to multiple
classes -- a limitation that we remove by considering semi-supervised
learning in Section \ref{sect:semi_supervised}. It should be noted
that we could have equivalently derived the CatGAN model starting from
the perspective of regularized information maximization (RIM) -- as
described in the appendix -- with an equivalent outcome.

\subsection{Problem setting}
\label{sect:problem}
As before, let $\mathcal{X} = \lbrace \bx^1, \dots \bx^N \rbrace$ be a
dataset of unlabeled examples. We consider the problem of
unsupervisedly learning a discriminative classifier $D$ from
$\mathcal{X}$, such that $D$ classifies the data into an a priori chosen
number of categories (or classes) $K$. Further, we require $D(\bx)$ to
give rise to a conditional probability distribution over categories;
that is $\sum_{k=1}^K p(y = k | \bx, D) = 1$. The goal of learning
then is to train a probabilistic classifier $D$ whose class
assignments satisfy some \emph{goodness of fit} measures. Notably,
since the true class distribution over examples is not known we have
to resort to an intermediary measure for judging classifier
performance, rather than just minimizing, e.g., the negative log
likelihood. Specifically, we will, in the following, always prefer $D$
for which the conditional class distribution $p(y | \bx, D)$ for a
given example $\bx$ has high certainty and for which
the marginal class distribution $p(y | D)$ is close to some prior
distribution $P(y)$ for all $k$. We will henceforth always assume a
uniform prior over classes, that is we expect that the amount of
examples per class in $\mathcal{X}$ is the same for all $k$:
$\forall k,k' \in K: p(y = k | D) = p(y = k' | D)$~\footnote{We
  discuss the possibility of using different priors in our framework
  in the appendix of this paper.}

A first observation about this problem is that it can naturally be
considered as a ``soft'' or probabilistic cluster assignment task. It
could thus, in principle, be solved by probabilistic clustering
algorithms such as regularized information maximization (RIM)
\citep{Krause_2010}, or the related entropy minimization
\citep{Grandvalet_2004}, or the early work on unsupervised
classification with phantom targets by \citet{Bridle_91}. All of these
methods are prone to overfitting to spurious correlations in the data
\footnote{In preliminary experiments we noticed that the MNIST dataset
  can, for example, be nicely separated into ten classes by creating
  2-3 classes for common noise patterns and collapsing together
  several ``real'' classes.}, a problem that we aim to mitigate by
pairing the discriminator with an adversarial generative model to
whose examples it must become robust. We note in passing, that our
method can be understood as a robust extension of RIM -- in which the
adversary provides an adaptive regularization mechanism. This
relationship is made explicit in Section
\ref{sect:relation_catgan_rim} in the appendix.

A somewhat obvious, yet important, second observation that can be made
is that the standard GAN objective cannot directly be used to solve
the described problem.  The reason for this is that while optimization
of Equation \eqref{eq:gan_loss} does result in a discriminative
classifier $D$ -- which must capture the statistics of the provided
input data -- this classifier is only useful for determining whether
or not a given example $\bx$ belongs to $\mathcal{X}$. In principle, we
could hope that a classifier which can model the data distribution
might also learn a feature representation (e.g. in case of neural
networks the hidden representation in the last layer of $D$) useful
for extracting classes in a second step; for example via
discriminative clustering.  It is, however, instructive to realize
that the means by which the function $D$ performs the binary
classification task -- of discriminating real from fake examples --
are not restricted in the GAN framework and hence the classifier will
focus mainly on input features which are not yet correctly modeled by
the generator. In turn, these features need not necessarily align with
our concept of classes into which we want to separate the data. They
could, in the worst case, be detecting noise in the data that stems
from the generator.

Despite these issues there is a principled, yet simple, way of
extending the GAN framework such that the discriminator can be used
for multi-class classification. To motivate this, let us consider
a change in protocol to the two player game behind the GAN framework (which
we will formalize in the next section): 
Instead of asking $D$ to predict the probability of $\bx$ belonging to
$\mathcal{X}$ we can require $D$ to assign all examples to one of $K$
categories (or classes), while staying uncertain of class assignments for
samples from the generative model $G$ -- which we expect will help make
the classifier robust. 
Analogously, we can change the problem posed to the generator from
``generate samples that belong to the dataset'' to ``generate samples
that belong to precisely one out of $K$ classes''.

If we succeeded at training such a classifier-generator pair --
and simultaneously ensured that the discovered $K$ classes
coincide with the classification problem we are interested in
(e.g. $D$ satisfies the \emph{goodness of fit} criteria outlined above) -- we
would have a general purpose formulation for training a classifier
from unlabeled data.

\begin{figure}
  \centering
  \includegraphics[width=4cm]{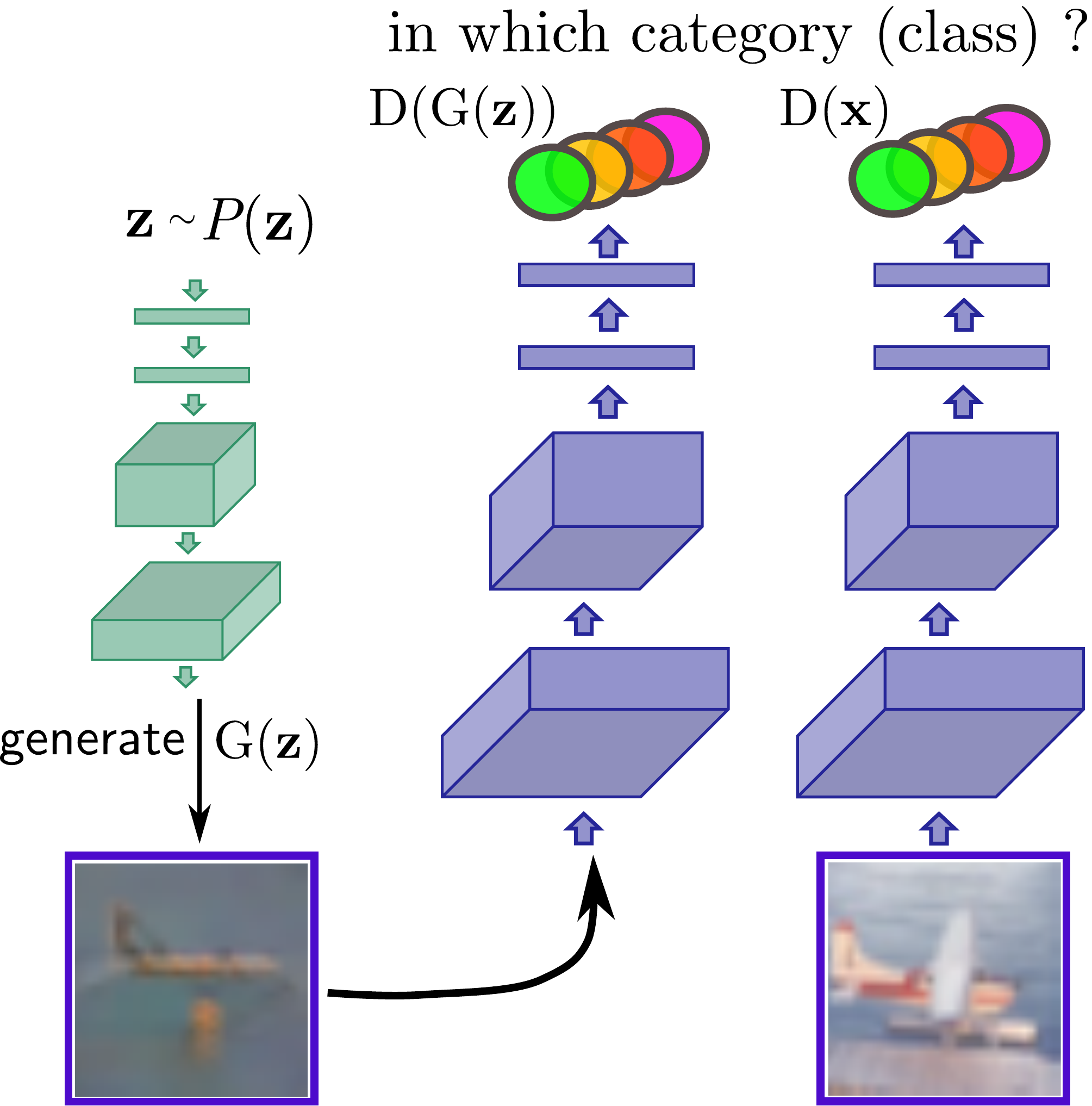} 
  \quad \quad
  \vline
  \quad \quad
    \includegraphics[height=4cm]{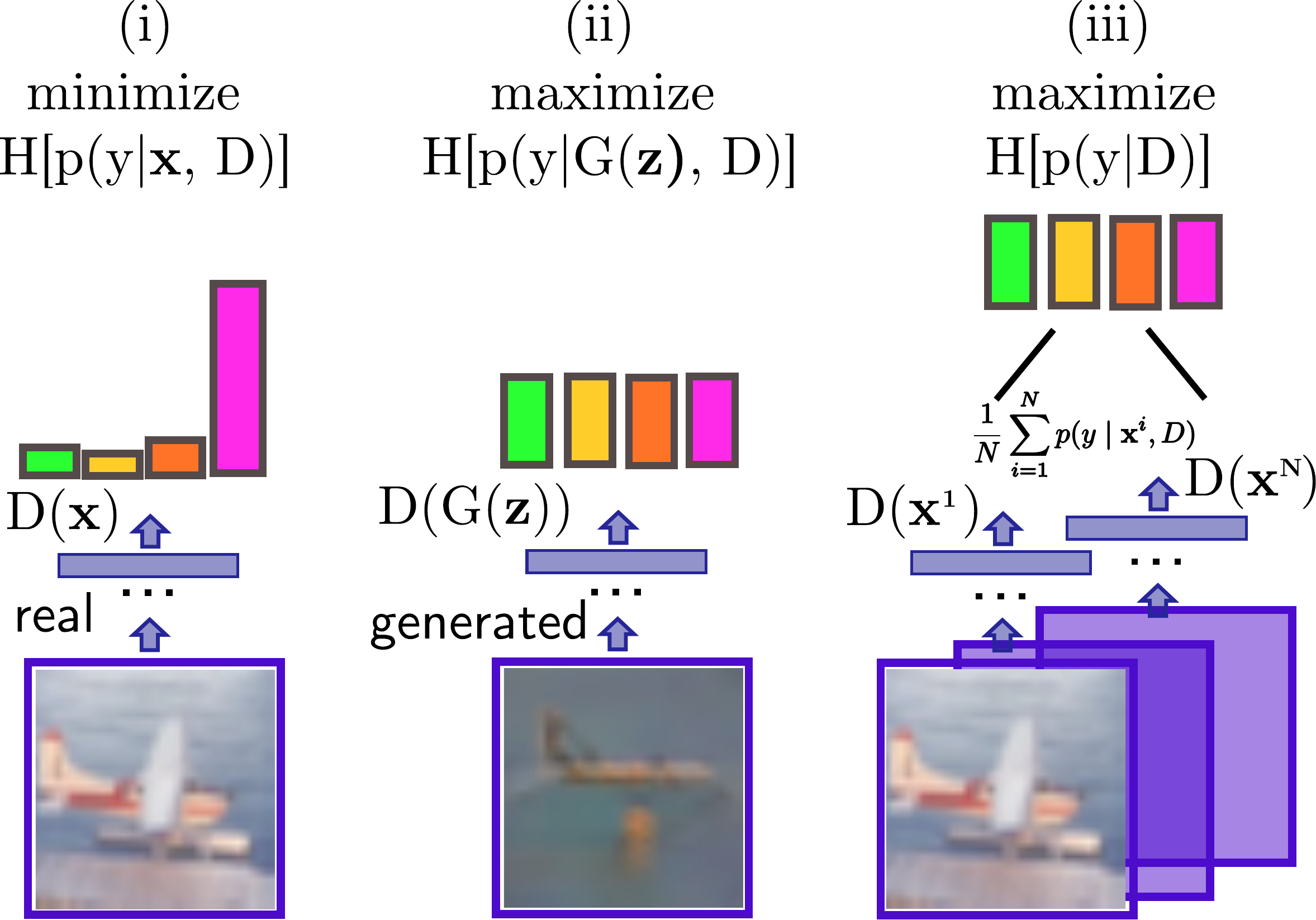}
    \caption{Visualization of the information flow through the
      generator (in green) and discriminator (in violet) neural networks (left). A sketch
      of the three parts (i) - (iii) of the objective function $\cLD$ for the
      discriminator (right). To obtain certain predictions the
      discriminator minimizes the entropy of $p(y | \bx, D)$, leading
      to a peaked conditional class distribution. To obtain uncertain
      predictions for generated samples the the entropy of $p(y |
      G(\bz), D)$ is maximized which, in the limit, would result in a uniform
      distribution. Finally, maximizing the marginal class entropy
      over all data-points leads to uniform usage of all classes.}
  \label{fig:reqs}
\end{figure}

\subsection{CatGAN objective}
\label{sect:objective}
As outlined above, the optimization problem that we want to solve
differs from the standard GAN formulation from Eq. \eqref{eq:gan_loss}
in one key aspect: instead of learning a binary discriminative
function, we aim to learn a discriminator that separates the data into
$K$ categories by assigning a label $y$ to each example
$\bx$. Formally, we define the discriminator $D(\bx)$ for this setting
as a differentiable function predicting logits for $K$ classes:
$D(\bx) \in \mathbb{R}^K$. The probability of example $\bx$ belonging
to one of the $K$ mutually exclusive classes is then given through a
softmax assignment based on the discriminator output:
\begin{equation}
  p(y = k | \bx, D) = \frac{e^{D_k(\bx)}}{\sum^K_{k=1} e^{D_k(\bx)}}.
\end{equation}

As in the standard GAN formulation we define the generator $G(\bz)$ to
be a function mapping random noise $\bz \in \mathbb{R}^Z$ to generated 
samples $\tilde{\bx} \in \mathbb{R}^I$:
\begin{equation}
  \tilde{\bx} = G(\bz), \ \text{with} \ z \sim P(\bz),
\end{equation}
where $P(\bz)$ again denotes an arbitrary noise distribution. For the
purpose of this paper both $D$ and $G$ are always parameterized as multi-layer
neural networks with either linear or sigmoid output. 

As informally described in Section \ref{sect:problem}, the
\emph{goodness of fit} criteria -- in combination 
with the idea that we want to use a generative model to regularize our
classifier -- directly dictate three requirements that a learned
discriminator should fulfill, and two requirements that the generator
should fulfill. We repeat these here before turning them into a
learnable objective function (a visualization of the requirements is shown
in Figure \ref{fig:reqs}). 

\textbf{Discriminator perspective.} The requirements to the
discriminator are that it should (i) be \emph{certain} of class
assignment for samples from $\mathcal{D}$, (ii) be \emph{uncertain} of
assignment for generated samples, and (iii) use all classes 
\emph{equally}~\footnote{Since we assume a uniform prior $P(y)$
  over classes.}.

\textbf{Generator perspective.} The requirements to the
generator are that it should (i) generate samples with highly
\emph{certain} class assignments, and (ii) equally distribute samples
across all $K$ classes.

We will now address each of these requirements in turn -- framing them
as maximization or minimization problems of class probabilities --
beginning with the perspective of the discriminator. Note
that without additional (label) information about the $K$ classes we
cannot directly specify which class probability $p(y = k | \bx, D )$
should be maximized to meet requirement (i) for any given $\bx$.  We can,
nonetheless, formally capture the intuition behind this requirement
through information theoretic measures on the predicted class
distribution.  The most direct measure that can be applied to this
problem is the Shannon entropy $H$, which is defined as the expected
value of the information carried by a sample from a given
distribution. Intuitively, if we want the class distribution $p(y \mid
\bx, D)$ conditioned on example $\bx$ to be highly peaked -- i.e. $D$
should be certain of the class assignment -- we want the
information content $H[ p(y \mid \bx, D) ]$
of a sample from it to be low,
since any draw from said distribution should almost always result in
the same class. If we, on 
the other hand, want the conditional class distribution to be flat (highly
uncertain) for examples that do not belong to $\mathcal{X}$ -- but
instead come from the generator --
we can maximize the entropy $H[ p(y \mid G(\bz), D) ]$, which, at the
optimum, will result in a uniform conditional distribution over
classes and fulfill requirement (ii).
Concretely, we can define the empirical estimate of the conditional
entropy over examples from $\mathcal{X}$ as
\begin{equation}
\begin{aligned}
  \mathbb{E}_{\bx \sim \mathcal{X}} \Big[ H \big[ p(y \mid \bx, D)
  \big ] \Big] &=
  \frac{1}{N} \sum^N_{i=1} H \big[ p(y \mid \bx^{i}, D) \big] \\
  &= \frac{1}{N} \sum^N_{i=1} - \sum^K_{k=1} p(y = k \mid \bx^i, D) \log
  p(y = k \mid \bx^i, D).
\label{eq:ent_x}
\end{aligned}
\end{equation}
The empirical estimate of the conditional entropy over samples from
the generator can be expressed as the expectation of $H[ p(y \mid
G(\bz), D) ]$ over the prior distribution $P(\bz)$ for the noise
vectors $\bz$, which we can further approximate through Monte-Carlo
sampling yielding
\begin{equation}
  \mathbb{E}_{\bz \sim P(\bz)} \Big[ H \big[ p(y \mid D(\bz), D) \big] \Big] \approx
  \frac{1}{M} \sum^M_{i=1} H \big[ p(y \mid G(\bz^i), D) \big], \
  \text{with } \ \bz^i \sim P(\bz),
\label{eq:ent_xrec}
\end{equation}
and where $M$ denotes the number of independently drawn samples (which
we simply set equal to $N$).
To meet the third requirement that all classes should be used equally --
corresponding to a uniform marginal distribution -- we can
maximize the entropy of the marginal class distribution as measured
empirically based on $\mathcal{X}$ and samples from $G$:
\begin{equation}
\begin{aligned}
  H_\mathcal{X} \Big [ p(y \mid D) \Big] &=
  H \Big [ \frac{1}{N} \sum^N_{i=1} p(y \mid \bx^i, D) \Big ], \\
  H_{G} \Big [ p(y \mid D) \Big] &\approx
  H \Big [ \frac{1}{M} \sum^M_{i=1} p(y \mid G(\bz^i), D) \Big ], \
  \text{with } \ \bz^i \sim P(\bz).
\label{eq:class_ent}
\end{aligned}
\end{equation}
The second of these entropies can readily be used to define the
maximization problem that needs to be satisfied for the requirement (ii)
imposed on the generator. Satisfying the condition (i) from
the generator perspective then finally amounts to minimizing rather
than maximizing Equation \eqref{eq:ent_xrec}.

Combining the definition from Equations
(\ref{eq:ent_x},\ref{eq:ent_xrec},\ref{eq:class_ent}) we can
define the CatGAN objective for the discriminator, which we refer to
with $\mathcal{L}_{D}$, and for the generator, which we refer to with
$\mathcal{L}_{G}$ as
\begin{equation}
\begin{aligned}
  \mathcal{L}_{D} &= \max_{D} \ H_{\mathcal{X}} \Big [ p( y \mid D ) \Big ] -  \mathbb{E}_{\bx \sim
                                \mathcal{X}} \Big[ H \big[ p(y
                                 \mid \bx, D) \big]  \Big] + \mathbb{E}_{\bz
    \sim P(\bz)} \Big[ H\big[ p(y \mid G(\bz), D) \big]  \Big],
 \\
  \mathcal{L}_{G} &= \min_{G} - H_{G} \Big [ p\big( y \mid D \big) \Big ] + 
  \mathbb{E}_{\bz \sim P(\bz)} \Big[ H \big[ p(y \mid G(\bz), D) \big] \Big],
\label{eq:objective_catgan}
\end{aligned}
\end{equation}
where $H$ denotes the empirical entropy as defined above and we chose
to define the objective for the generator $\mathcal{L}_{G}$ as a
minimization problem to make the analogy to Equation
\eqref{eq:gan_loss} apparent. This formulation satisfies all
requirements outlined above and has a simple information theoretic
interpretation: Taken together the first two terms in
$\mathcal{L}_{D}$ are an estimate of the mutual information between
the data distribution and the predicted class distribution -- which
the discriminator wants to maximize while minimizing information it
encodes about $G(\bz)$. Analogously, the first two terms in
$\mathcal{L}_{G}$ estimate the mutual information between the
distribution of generated samples and the predicted class
distribution.

Since we are interested in optimizing the objectives from Equation
\eqref{eq:objective_catgan} on large datasets we would like both
$\cLG$ and $\cLD$ to be amenable to to optimization via mini-batch
stochastic gradient descent on batches $\mathcal{X}_B$ of data -- with
size $B \ll N$ -- drawn independently from $\mathcal{X}$. The conditional entropy terms in
Equation \eqref{eq:objective_catgan} both only consist of sums over
per example entropies, and can thus trivially be adapted for batch-wise
computation. The marginal entropies $H_{\mathcal{X}} [ p( y \mid
D ) ]$ and $H_{G} [ p\big( y \mid D \big) ]$, however,
contain sums either over the whole dataset $\mathcal{X}$ or over a
large set of samples from $G$ \emph{within} the entropy
calculation and therefore cannot be split into ``per-batch'' terms. 
If the number of categories $K$ that the discriminator needs
to predict is much smaller than the batch size $B$, a simple fix to
this problem is to estimate the marginal class distributions over the
$B$ examples in the random mini-batch only: $H_\mathcal{X} [ p(y \mid
D) ] \approx H [ \frac{1}{B} \sum_{\bx \in \mathcal{X}_B} p(y \mid
\bx^i, D) ]$.
For $H_{G} [ p( y \mid D ) ]$ we can, similarly, calculate an estimate
using $B$ samples only -- instead of using $M = N$ samples. We note
that while this approximation is reasonable for the problems we
consider (for which $K <= 10$ and $B = 100$) it will be problematic for scenarios in
which we expect a large number of categories. In such a setting one
would have to estimate the marginal class distribution over multiple batches
(or periodically evaluate it on a larger number of examples).

\subsection{Extension to semi-supervised learning}
\label{sect:semi_supervised}
We will now consider adapting the formulation from Section
\ref{sect:objective} to the semi-supervised setting. Let
$\mathcal{X}^L = \lbrace (\bx^1, \by^1), (\bx^L, \by^L) \rbrace$ be a
set of L labeled examples, with label vectors $\by^i \in \mathbb{R}^K$
in one-hot encoding, that are provided in addition to the $N$
unlabeled examples contained in $\mathcal{X}$. These additional
examples can be incorporated into the objectives from Equation
\eqref{eq:objective_catgan} by calculating a cross-entropy term
between the predicted conditional distribution $p(y \mid \bx, D)$ and
the true label distribution of examples from $\mathcal{X}^L$ (instead
of the entropy term $H$ used for unlabeled examples). The
cross-entropy for a labeled data pair $(\bx, \by)$ is given as
\begin{equation}
      CE \big[ \by, p(y \mid \bx, D) \big] = - \sum^K_{i=1} y_i \log p(y = y_i \mid \bx, D).
\end{equation}
The semi-supervised CatGAN problem is then given through the two
objectives $\mathcal{L}^L_{D}$ (for the discriminator) and $\mathcal{L}^L_{G}$ (for the generator) with
\begin{equation}
\begin{aligned}
  \mathcal{L}^L_{D} = \max_{D} \ &H_{\mathcal{X}} \Big [ p( y \mid D ) \Big ] -  \mathbb{E}_{\bx \sim
                                \mathcal{X}} \Big[ H \big[ p(y
                                 \mid \bx, D) \big]  \Big] + \mathbb{E}_{\bz
    \sim P(\bz)} \Big[ H\big[ p(y \mid G(\bz), D) \big]  \Big] \\
   + &\lambda \mathbb{E}_{(\bx, \by) \sim \mathcal{X}^L} \Big[ CE \big[ \by,  p(y \mid \bx, D) \big]  \Big],
\label{eq:objective_ss_catgan}
\end{aligned}
\end{equation}
where $\lambda$ is a cost weighting term and where $\mathcal{L}^L_{G}$
is the same as in Equation \eqref{eq:objective_catgan}:
$\mathcal{L}^L_{G} = \mathcal{L}_{G}$.

\subsection{Implementation Details}
In our experiments both the generator and the discriminator are always
parameterized through neural networks. The details of architectural
choices for each considered benchmark are given in the appendix, while
we only cover major design choices in this section. 

GANs are known to be hard to train due to several unfortunate
circumstances. First, the formulation from Equation
\eqref{eq:gan_loss} can become unstable if the discriminator learns
too quickly (in which case the loss for the generator saturates). Second, the
generator might get stuck generating one mode of the
data 
or it may start wildly switching between generating different modes during
training.

We therefore take two measures to stabilize training. First, we use
batch normalization \citep{Ioffe_2015} in all layers of the
discriminator and all but the last layer (the layer producing
generated examples $\tilde{\bx}$) of the generator. This helps bound
the activations in each layer and we empirically found it to prevent
mode switching of the generator as well as to increase generalization
capabilities of the discriminator in the few labels
case. Additionally, we regularize the discriminator by applying noise
to its hidden layers. While we did find dropout
\citep{Hinton_arxiv2012} to be effective for this purpose, we found
Gaussian noise added to the batch normalized hidden activations to
yield slightly better performance.
We suspect that this is mainly due to the fact that dropout noise can
severely affect mean and variance computation during
batch-normalization -- whereas Gaussian noise on the activations for
which to compute these statistics is a natural assumption.

\section{Empirical Evaluation}
The results of our empirical evaluation are given in Tables
\ref{tab:pi_mnist}, \ref{tab:cnn_mnist} and \ref{tab:cifar10}. As can
be seen, our method is competitive to the state of the art on almost
all datasets. It is only slightly outperformed by the Ladder network
utilizing denoising costs in each layer of the neural network.

\subsection{Clustering with CatGANs}
Since categorization of unlabeled data is inherently linked to
clustering we performed a first set of experiments on common synthetic
datasets that are often used to evaluate clustering algorithms. We
compare the CatGAN algorithm with standard k-means clustering and RIM
with neural networks as discriminative models, which amounts to
removing the generator from the CatGAN model and adding $\ell_2$
regularization (see Section \ref{sect:relation_catgan_rim} in the
appendix for an explanation). We considered three standard synthetic
datasets -- with feature dimensionality two, thus
$\bx \in \mathbb{R}^2$ -- for which we assumed the optimal number of
clusters $K$ do be known: the ``two moons'' dataset (which contains
two clusters), the ``circles'' arrangement (again containing two
clusters) and a simple dataset with three isotropic Gaussian blobs of
data. 

\begin{figure}[t]
  \includegraphics[width=14cm]{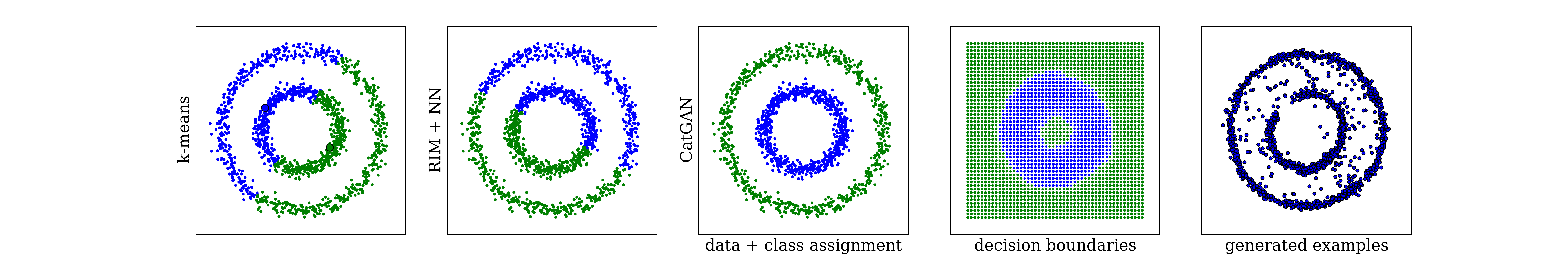}
  \caption{Comparison between k-means (left), RIM (middle) and CatGAN
    (rightmost three) -- with neural networks -- on the ``circles''
    dataset with $K =2$. Blue and green denote class assignments to
    the two different classes. For CatGAN we visualize class
    assignments -- both on the dataset and on a larger region of the
    input domain -- and generated samples. Best viewed in color.}
  \label{fig:circles_main}
\end{figure}

In Figure \ref{fig:circles_main} we show the results of that
experiment for the ``circles'' dataset (plots for the other two
experiments are relegated to Figures
\ref{fig:blobs_compare}-\ref{fig:circles_compare} in the appendix due
to space constraints). In summary, the simple clustering assignment
with three data blobs is solved by all algorithms. For the two more
difficult examples both k-means and RIM fail to ``correctly'' identify
the clusters: (1) k-means fails due to the euclidean distance measure it
employs to evaluate distances between data points and cluster
centers, (2) in RIM the objective function only specifies that the deep
network has to separate the data into two equal classes, without any
geometric constraints~\footnote{We tried to rectify this by adding
  regularization (we tried both $\ell_2$ regularization and adding
  Gaussian noise) but that did not yield any improvement}. In the
CatGAN model, on the other hand, the discriminator has to place its
decision boundaries such that it can easily detect a non-optimal
adversarial generator which seems to coincide with the correct cluster
assignment. Additionally, the generator quickly learns to generate the
datasets in all cases.

\begin{table}[t]
  \begin{tabular}{lccc}
    \multirow{2}{*}{\bf Algorithm} & \multicolumn{3}{c}{PI-MNIST
                                     test error (\%) with $n$ labeled examples} \\
    & $n = 100$ & $n = 1000$ & All  \\
    \hline
    MTC \citep{Rifai_2011} & 12.03 & 100 & 0.81 \\
    PEA \citep{Bachman_NIPS2014} & 10.79 & 2.33 & 1.08 \\
    PEA+ \citep{Bachman_NIPS2014} & 5.21 & 2.67 & -  \\
    VAE+SVM \citep{Kingma_NIPS2014} & 11.82 ($\pm$ 0.25) & 4.24 ($\pm$ 0.07) & - \\
    SS-VAE \citep{Kingma_NIPS2014} & 3.33 ($\pm$ 0.14) & 2.4 ($\pm$ 0.02) & 0.96
    \\
    Ladder $\Gamma$-model \citep{Rasmus_NIPS2015} & 4.34 ($\pm$ 2.31) & 1.71 ($\pm$ 0.07) & 0.79 ($\pm$ 0.05) \\
    Ladder full \citep{Rasmus_NIPS2015} & \textbf{1.13 ($\pm$ 0.04)} & \textbf{1.00 ($\pm$ 0.06)} & - \\
    \hline
    RIM + NN & 16.19 ($\pm$ 3.45)  & 10.41 ($\pm$ 0.89) \\ 
    GAN + SVM & 28.71 ($\pm$ 7.41) & 13.21 ($\pm$ 1.28) \\
    CatGAN (\emph{unsupervised}) & 9.7 & &  \\
    CatGAN (semi-supervised) & 1.91 ($\pm$ 0.1) & 1.73 ($\pm$ 0.18) & 0.91  \\  
  \end{tabular}
  \caption{Classification error, in percent, for the permutation invariant MNIST problem with a reduced number of labels. Results are averaged over 10 different sets of labeled examples.}
  \label{tab:pi_mnist}
\end{table}

\subsection{Unsupervised  and semi-supervised learning of image features}
We next evaluate the capabilities of the CatGAN model on two
image recognition datasets. We performed
experiments using fully connected and convolutional
networks on MNIST~\citep{LeCun_NC1989} and CIFAR-10~\citep{Krizhevsky2009}. We either used the full set of
labeled examples or a reduced set of labeled examples and kept
the remaining examples for semi-supervised or unsupervised learning.

We performed experiments using two setups: (1) using a subset of
labeled examples we optimized the semi-supervised objective from
Equation \eqref{eq:objective_catgan}, and (2) using \emph{no labeled
  examples} we optimized the unsupervised objective from Equation
\eqref{eq:objective_ss_catgan} with $K = 20$ ``pseudo'' categories. In
setup (2) learning was followed by a category matching step. In this
second step we simply looked at 100 examples from a validation set (we
always kept $10000$ examples from the training set for validation) for
which we assume the correct labeling to be known, and assigned each
pseudo category $y_k$ to be indicative of one of the true classes
$c_i \in \lbrace 1 \dots 10 \rbrace$. Specifically we assign $y_k$ to
the class $i$ for which the count of examples that were classified as
$y_k$ and belonged to $c_i$ was maximal. This setup hence bears some
similarity to one-shot learning approaches from the literature (see
e.g. \citet{FeiFeiFergusPeronaPAMI} for an application to computer
vision). Since no learning is involved in the actual matching step we
-- somewhat colloquially -- refer to this setup as \emph{half-shot
  learning}.

The results for the experiment on the permutation invariant MNIST
(PI-MNIST) task are listed in Table \ref{tab:pi_mnist}. The table also
lists state-of-the-art results for this benchmark as well as two
baselines: a version of our algorithm where the generator is removed
-- but all other pieces stay in place -- which we call RIM + NN due to
the relationship between our algorithm and RIM; and the discriminator
stemming from a standard GAN paired with an SVM trained based on
features from it~\footnote{Specifically, we first train a
  generator-discriminator using the standard GAN objective and then
  extract the last layer features from the discriminator on the
  available labeled examples, and use them to train an SVM.}.

While both the RIM and GAN training objectives do produce features
that are useful for classifying digits, their performance is far worse
than the best published result for this setting. The semi-supervised
CatGAN, on the other hand, comes close to the best results, works
remarkably well even with only 100 labeled examples, and is only
outperformed by the Ladder network with a specially designed denoising
objective in each layer.
Perhaps more surprisingly the half-shot learning procedure
described above results in a classifier that achieves $9.7 \%$
error \emph{without the need for any label information} during training. 

Finally, we performed experiments with convolutional discriminator
networks and de-convolutional~\citep{Zeiler_ICCV2011} generator
networks (using the same up-sampling procedure from
\citet{DosSprChair15}) on MNIST and CIFAR-10. As before, details on
the network architectures are given in the appendix. The results are
given in Table \ref{tab:cnn_mnist} and \ref{tab:cifar10} and are
qualitatively similar to the PI-MNIST results; notably the
unsupervised CatGAN again performs very well, achieving a classification error
of $4.27$. The discriminator trained with the semi-supervised CatGAN
objective performed well on both tasks, matching the state of the art
on CIFAR-10 with reduced labels.

\begin{table}
  \begin{tabular}{lcc}
    \multirow{2}{*}{\bf Algorithm} & \multicolumn{2}{c}{MNIST test error (\%) with $n$ labeled examples} \\
                                   & $n = 100$ & All \\
    \hline
    EmbedCNN \citep{Weston_2012} & 7.75 & - \\
    SWWAE \citep{Zhao_2015} & 8.71 $\pm 0.34$ & 0.71 \\
    Small-CNN \citep{Rasmus_NIPS2015} & 6.43 ($\pm$ 0.84) & 0.36 \\
    Conv-Ladder $\Gamma$-model \citep{Rasmus_NIPS2015} & \textbf{0.86 ($\pm$ 0.41)} & - \\
    \hline
    RIM + CNN & 10.75 ($\pm$ 2.25) & 0.53  \\
    Conv-GAN + SVM & 15.43 ($\pm$ 1.72) & 9.64 \\
    Conv-CatGAN (\emph{unsupervised}) & 4.27  &  \\
    Conv-CatGAN (semi-supervised) & 1.39 ($\pm$ 0.28) & 0.48 \\
  \end{tabular}
  \caption{Classification error, in percent, for different learning methods in combination with convolutional neural networks (CNNs) with a reduced number of labels.}
  \label{tab:cnn_mnist}
\end{table}

\begin{table}
  \begin{tabular}{lcc}
    \multirow{2}{*}{\bf Algorithm} & \multicolumn{2}{c}{CIFAR-10 test error (\%) with $n$ labeled examples} \\
                                   & $n = 4000$ & All \\
   \hline
    View-Invariant k-means \cite{Yuhui_2013} & 27.4 ($\pm$ 0.7) & 18.1 \\
    Exemplar-CNN \citep{Dosovits_NIPS2014} & 23.4 ($\pm$ 0.2) & 15.7 \\
    Conv-Ladder $\Gamma$-model \citep{Rasmus_NIPS2015} & \textbf{20.09 ($\pm$ 0.46)} & \textbf{9.27} \\
    \hline
    Conv-CatGAN (semi-supervised) & \textbf{19.58 ($\pm$ 0.58)} & 9.38 \\
  \end{tabular}
  \caption{Classification error for different methods on the CIFAR-10 dataset (without data augmentation) for the full dataset and a reduced set of 400 labeled examples per class.}
  \label{tab:cifar10}
\end{table}

\subsection{Evaluation of the generative model}
\begin{figure}[t]
  \centering
  \includegraphics[width=0.47\textwidth]{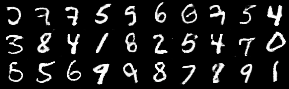}
  \includegraphics[width=0.45\textwidth]{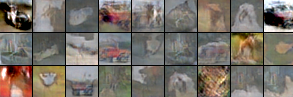}
  \caption{Exemplary images produced by a generator trained using the
    semi-supervised CatGAN objective. We show samples for a generator
    trained on MNIST (left) CIFAR-10 (right).}
  \label{fig:samples_lfw}
\end{figure}

Finally, we qualitatively evaluate the capabilities of the generative model. We
trained an unsupervised CatGAN on MNIST,
LFW and CIFAR-10 and plot samples generated by these models in Figure
\ref{fig:samples_lfw}. As an additional quantitative evaluation we
compared the unsupervised CatGAN model trained on MNIST with other
generative models based on the log likelihood of generated samples (as
measured through a Parzen-window estimator). The full results of this
evaluation are given in Table \ref{tab:ll_mnist} in the appendix. In
brief: The CatGAN model performs comparable to the best existing
algorithms, achieving a log-likelihood of $237 \pm 6$ on MNIST; in
comparison, \citet{Goodfellow_NIPS2014} report $225 \pm 2$ for GANs.
We note, however, that this does not necessarily mean that the CatGAN
model is superior as comparing generative models with respect to
log-likelihood measured by a Parzen-window estimate can be misleading
(see \cite{Theis_2015} for a recent in-depth discussion).

\section{Relation to prior work}
\label{sect:prior_work}
As highlighted in the introduction our method is related to, and
stands on the shoulders of, a large body of literature on unsupervised
and semi-supervised category discovery with machine learning
methods. While a comprehensive review of these methods is out of the
scope for this paper we want to point out a few interesting connections. 

First, as already discussed, the idea of minimizing entropy of a
classifier on unlabeled data has been considered several times already
in the literature~\citep{Bridle_91,Grandvalet_2004,Krause_2010}, and
our objective function falls back to the regularized information
maximization from \citet{Krause_2010} when the generator is removed
and the classifier is additionally $\ell_2$ regularized\footnote{We
  note that we did not find $\ell_2$ regularization to help in our
  experiments.}. Several researchers have recently also reported
successes for unsupervised learning with pseudo-tasks, such as
self-supervised labeling a set of unlabeled training examples
\citep{Lee_ICML_WCLR2013}, learning to recognize pseudo-classes obtained through
data augmentation \citep{Dosovits_NIPS2014} and learning with
pseudo-ensembles~\citep{Bachman_NIPS2014}, in which a set of models
(with shared parameters) are trained such they agree on their
predictions, as measured through e.g. cross-entropy. While on first
glance these appear only weakly related, they are strongly connected to
entropy minimization as, for example, concisely explained in~\citet{Bachman_NIPS2014}.  

From the generative modeling perspective, our model is a direct
descendant of the generative adversarial networks
framework~\citep{Goodfellow_NIPS2014}. Several extensions to this
framework have been developed recently, including conditioning on a
set of variables~\citep{Gauthier_2014,Mirza_2014} and hierarchical generation using
Laplacian pyramids~\citep{Denton_NIPS2015}. These are orthogonal to
the methods developed in this paper and a combination of, for example,
CatGANs with more advanced generator architectures is an interesting
avenue for future work.

\section{Conclusion}
We have presented categorical generative adversarial networks, a framework for robust unsupervised and semi-supervised learning. Our method combines neural network classifiers with an adversarial generative model that regularizes a discriminatively trained classifier. We found the proposed method to yield classification performance that is competitive with state-of-the-art results for semi-supervised learning for image classification and further confirmed that the generator, which is learned alongside the classifier, is capable of generating images of high visual fidelity.

\subsubsection*{Acknowledgments}
The author would like to thank Alexey Dosovitskiy, Alec Radford, Manuel Watter, Joschka Boedecker and Martin Riedmiller for extremely helpful discussions on the contents of this manuscript. Further, huge thanks go to Alec Radford and the developers of Theano \citep{Theano_scipy,Theano_2012} and Lasagne \citep{Lasagne_2015} for sharing research code. This work was funded by the the German Research Foundation (DFG) within the priority program ``Autonomous learning'' (SPP1597).

\bibliography{catgan_iclr}
\bibliographystyle{iclr2016_conference}

\section*{Appendix}
\begin{appendix}

\section{On the relation between CatGAN and GAN}
In this section we will make the relation between the CatGAN objective
from Equation \eqref{eq:objective_catgan} in the main paper and the
GAN objective given by Equation \eqref{eq:gan_loss} more directl
apparent.  Starting from the CatGAN objective let us consider the case
$K=1$. In this case the conditional probabilities should model binary
dependent variables (and are thus no longer multinomial). The
``correct'' choice for the discriminative model is a logistic
classifier with output $D(\bx) \in \mathbb{R}$ with
conditional probability $p(y = 1 \mid \bx, D)$ given as
$p(y = 1 \mid \bx, D) = \frac{e^{D(\bx)}}{e^{D(\bx)} + 1} = \frac{1}{1 + e^{-D(\bx)}}$. Using this
definition The discriminator loss $\mathcal{L}_D$ from Equation
\eqref{eq:objective_catgan} can be expanded to give
\begin{equation}
\begin{aligned}
  \mathcal{L}^1_{D} = \max_{D} - &\mathbb{E}_{\bx \sim
                                \mathcal{X}} \Big[ H \big[ p(y
                                 \mid \bx, D) \big]  \Big] + \mathbb{E}_{\bz
    \sim P(\bz)} \Big[ H\big[ p(y \mid G(\bz), D) \big]  \Big]
 \\
 = \max_{D} \  &\mathbb{E}_{\bx \sim \mathcal{X}} \Big[  p_{\bx} \log  p_{\bx} + (1 - p_{\bx}) \log (1 - p_{\bx})  \Big]  \\
  + &\mathbb{E}_{\bz \sim P(\bz)} \Big[-p_{G(\bz)} \log p_{G(\bz)} - (1 - p_{G(\bz)}) \log (1 - p_{G(\bz)}) \Big],
\label{eq:objective_catgan_one}
\end{aligned}
\end{equation}
where we introduced the short notation
$p_{\bx} = p(y = 1 \mid \bx, D)$,
$p_{G(\bz)} = p(y = 1 \mid G(\bz), D)$ and dropped the entropy term
$H_{\mathcal{X}} \Big [ p( y \mid D ) \Big ]$ concerning the empirical
class distribution as we only consider one class and hence the classes
are equally distributed by definition. Equation \eqref{eq:objective_catgan_one}
now is similar to the GAN objective but pushes the conditional
probability for samples from $\mathcal{X}$ to $0$ or $1$ and the
probability for generated samples towards $0.5$. To obtain a
classifier which predicts $p(y=1 \mid \bx, D)$ we can replace the
entropy $H \big[ p(y \mid \bx, D) \big]$ with the cross-entropy $CE \big[1, p(y \mid \bx, D) \big]$ yielding
\begin{equation}
\begin{aligned}
  \mathcal{L}^1_{D} = \max_{D} \  &\mathbb{E}_{\bx \sim \mathcal{X}} \Big[  \log  p(y=1 \mid \bx, D)  \Big]  \\
  + &\mathbb{E}_{\bz \sim P(\bz)} \Big[-p_{G(\bz)} \log p_{G(\bz)} - (1 - p_{G(\bz)}) \log (1 - p_{G(\bz)}) \Big],
\label{eq:objective_catgan_one_2}
\end{aligned}
\end{equation}
which is equivalent to the discriminative part of the GAN formulation
except for the fact that optimization of Equation
\eqref{eq:objective_catgan_one_2} will result in examples from the
generator being pushed towards the decision boundary of
$p(y=1 \mid G(\bz), D) = 0.5$ rather than $p(y=1 \mid G(\bz), D) = 0$.  An
equivalent derivation can be made for the generator objective
$\mathcal{L}_G$ leading to a symmetric objective -- just as in the GAN
formulation.

\section{On the relation between CatGAN and RIM}
\label{sect:relation_catgan_rim}
In this section we re-derive CatGAN as an extension to the RIM
framework from \citet{Krause_2010}. As in the main paper we will
restrict ourselves to the unsupervised setting but an extension to the
semi-supervised setting is straight-forward. The idea behind RIM is to
train a discriminative classifier, which we will suggestively call
$D$, from unlabeled data. The objective that is maximized for this
purpose is the mutual information between the data distribution and
the predicted class labels, which can be formalized as
\begin{equation}
\begin{aligned}
  \mathcal{L}^{RIM} &= \max_{D} \ H_{\mathcal{X}} \Big [ p( y \mid D ) \Big ] - \mathbb{E}_{\bx \sim
                                \mathcal{X}} \Big[ H \big[ p(y
                                 \mid \bx, D) \big]  \Big] - \gamma R(D),
\label{eq:objective_rim}
\end{aligned}
\end{equation}
where the entropy terms are defined as in the main paper and $R(D)$ is
a regularization term acting on the discriminative model. In
\citet{Krause_2010} $D$ was chosen as a logistic regression classifier
and $R(D)$ consisted of $\ell_2$ regularization on the discriminator
weights.  If we instantiate $D$ to be a neural network we obtain the
baseline RIM + NN which we considered in our experiments.

To connect the RIM objective to the CatGAN formulation from Equation
\eqref{eq:objective_catgan} we can set let
$R(D) = - \mathbb{E}_{\bz \sim P(\bz)} \Big[ H\big[ p(y \mid G(\bz),
D) \big] \Big]$,
that is we let $R(D)$ measure the negative entropy of samples from the
generator. With this setting we achieve equivalence between
$\mathcal{L}^{RIM}$ and $\mathcal{L}_{D}$. If we now also train the
generator $G$ alongside the discriminator $D$ using the objective
$\mathcal{L}_{G}$ we arrive at the CatGAN formulation.

\section{On different priors for the empirical class distribution}
In the main paper we always assumed a uniform prior over classes, that
is we enforced that the amount of examples per class in $\mathcal{X}$
is the same for all $k$:
$\forall k,k' \in K: p(y = k' | D) = p(y = k' | D)$. This was achieved
by maximizing the entropy of the class distribution
$H_{\mathcal{X}} \Big [ p( y \mid D ) \Big ]$. If this prior
assumption is not valid our method could be extended to different
prior distributions $P(y)$ similar to how RIM can be adapted (see
Section 5.2 of \citet{Krause_2010}). This becomes easy to see ny
noticing the relationship between the Entropy and the KL divergence:
$H_{\mathcal{X}} \Big [ p( y \mid D ) \Big ] = \log(K) - KL(p( y \mid
D ) \| U) $
where $U$ denotes the discrete uniform distribution. We can thus
simply drop the constant term $\log(K)$ and use
$- KL(p( y \mid D ) \| \mathcal{U})$ directly, allowing us to replace
$\mathcal{U}$ with an arbitrary prior $P(y)$ -- as long as we can
differentiate through the computation of the KL divergence (or
estimate it via sampling). 

\section{Detailed explanation of the training procedure}
As mentioned in the main Paper we perform training by alternating
optimization steps on $\mathcal{L}_D$ and $\mathcal{L}_G$. More
specifically, we use batch size $B=100$ in all experiments and
approximate the expectations in Equation \eqref{eq:objective_catgan}
and Equation \eqref{eq:objective_ss_catgan} using $100$ random
examples from $\mathcal{X}$, $\mathcal{X}^L$ and the generator
$G(\bz)$ respectively. We then do one gradient ascent step on the
objective for the discriminator followed by one gradient descent step
on the objective for the generator. We also added noise to all layers
as mentioned in the main paper. Since adding noise to the network can
result in instabilities in the computation of the entropy terms from
our objective (due to small values inside the logarithms which are
multiplied with non-negative probabilities) we added noise only to the
terms not appearing inside logarithms. That is we effectively replace
$H[p(y \mid \bx, D)]$ with the cross-entropy
$CE[p(y \mid \bx, D), p(y \mid \bx, \hat{D})]$, where $\hat{D}$ is the
network with added noise and additionally truncate probabilities to be
bigger than $1e-4$.  During our evaluation we experimented with Adam
\citep{Kingma_2015} for adapting learning rates but settled for a
hybrid between \citep{Schaul2013pesky} and rmsprop
\citep{Tieleman2012}, called SMORMS3 \citep{Funk_2015} which we found
slightly easier to use as it only has one free parameter -- a maximum
learning rate -- which we did always set to $0.001$.

\subsection{Details on network architectures}
\subsubsection{Synthetic benchmarks}
For the synthetic benchmarks we used neural networks with three hidden
layers, containing 100 leaky rectified linear units each (leak rate
$0.1$), both for the discriminator and the generator (where
applicable). Batch normalization was used in all layers (with added
Gaussian noise with standard deviation $0.05$) and the dimensionality
of the noise vectors $\bz$ for the CatGAN model was chosen to be $10$
for. Note that while such large networks are most certainly an
``overkill'' for the considered benchmarks, we did chose these
settings to ensure that learning was easily possible. We also
experimented with smaller networks but did not find them to result in
better decision boundaries or more stable learning.

\begin{table}[h]
\caption{The discriminator and generator CNNs used for MNIST.}
\label{base-models}
\begin{center}
\begin{small}
\begin{tabular}{l|l}
\multicolumn{2}{c}{\textbf{Model}} \\
\bf discriminator $D$        & \bf generator $G$  \\
\hline
Input $28 \times 28$ Gray image & Input $\bz \in \mathbb{R}^{128}$ \\
\hline
$5 \times 5$ conv. $32$ lReLU & $8 \times 8 \times 96$ fc lReLU  \\
\hline
$3 \times 3$ max-pool, stride $2$ & $2 \times 2$ perforated up-sampling \\
\hline
$3 \times 3$ conv. $64$ lReLU & $5 \times 5$ conv. $64$ lReLU \\
$3 \times 3$ conv. $64$ lReLU & \\
\hline
$3 \times 3$ max-pool, stride $2$ & $2 \times 2$ perforated
                                    up-sampling \\
\hline
$3 \times 3$ conv. $128$ lReLU & $5 \times 5$ conv. $64$ lReLU \\
$1 \times 1$ conv. $10$ lReLU & $5 \times 5$ conv. $1$ lReLU \\
$128$ fc lReLU & \\ 
10-way softmax & \\
\end{tabular}
\end{small}
\end{center}
\end{table}

\begin{table}[h]
\caption{The discriminator and generator CNNs used for CIFAR-10.}
\label{base-models-cifar}
\begin{center}
\begin{small}
\begin{tabular}{l|l}
\multicolumn{2}{c}{\textbf{Model}} \\
\bf generator $G$        & \bf discriminator $D$  \\
\hline
Input $32 \times 32$ RGB image & Input $\bz \in \mathbb{R}^{128}$ \\
\hline
$3 \times 3$ conv. $96$ lReLU & $8 \times 8 \times 192$ fc lReLU  \\
$3 \times 3$ conv. $96$ lReLU &  \\
$3 \times 3$ conv. $96$ lReLU &  \\
\hline
$2 \times 2$ max-pool, stride $2$ & $2 \times 2$ perforated up-sampling \\
\hline
$3 \times 3$ conv. $192$ lReLU & $5 \times 5$ conv. $96$ lReLU \\
$3 \times 3$ conv. $192$ lReLU & $5 \times 5$ conv. $96$ lReLU \\
$3 \times 3$ conv. $192$ lReLU & \\
\hline
$3 \times 3$ max-pool, stride $2$ & $2 \times 2$ perforated
                                    up-sampling \\
\hline
$3 \times 3$ conv. $192$ lReLU & $5 \times 5$ conv. $96$ lReLU \\
$1 \times 1$ conv. $192$ lReLU & \\ 
$1 \times 1$ conv. $10$ lReLU & $5 \times 5$ conv. $1$ lReLU \\
global average & \\
10-way softmax & \\
\end{tabular}
\end{small}
\end{center}
\end{table}

\subsubsection{Permutation invariant MNIST}
For the permutation invariant MNIST task we used fully connected
generator and discriminator networks with leaky rectified linearities
(and a leak rate of $0.1$). For the discriminator we used the same
architecture as in \citet{Rasmus_NIPS2015}, consisting of a network
with 5 hidden layers (with sizes 1000, 500, 250, 250, 250
respectively). Batch normalization was applied to each of these layers
and Gaussian noise was added to the batch normalized responses as well
as the pixels of the input images (with a standard deviation of 0.3).
The generator for this task consisted of a network with three hidden
layers (with hidden sizes 500, 500, 1000) respectively. The output of
this network was of size $784 = 28 \times 28$, producing pixel images,
and used a sigmoid nonlinearity. The noise dimensionality for vectors
$\bz$ was chosen as $Z = 128$ and the cost weighting factor $\lambda$
was simply set to $\lambda = 1$. Note that on MNIST the classifier
quickly learns to classify the few labeled examples leading to a
vanishing supervised cost term; in a sense the labeled examples serve
more as a ``class initialization'' in these experiments. We note that we
found many different architectures to work well for this benchmark and
merely settled on the described settings to keep our results somewhat
comparable to the results from \citet{Rasmus_NIPS2015}.

\subsubsection{CNNs for MNIST and CIFAR-10}
Full details regarding the CNN architectures used both for the
generator and the discriminator are given in Table \ref{base-models}
for MNIST and in Table \ref{base-models-cifar} for CIFAR-10. They are
similar to the models from \citet{Rasmus_NIPS2015} who, in turn,
derived them from the best models found by \citet{SprDosBroRied15}. In
the Table ReLU denotes rectified linear units, lReLU denotes leaky
rectified linear units (with leak rate 0.1), fc stands for a fully
connected layer, conv for a convolutional layer and perforated
up-sampling denotes the deconvolution approach derived in
\citet{DosSprChair15} and \citet{OseSoySma2014}. 
\TODO{Double check that these network definitions are correct}

\section{Additional experiments}

\subsection{Quantitative evaluation of the generative model}
Table \ref{tab:ll_mnist} shows the sample log-likelihood for samples
from an unsupervised CatGAN model.  The CatGAN model performs
comparable to the best existing algorithms; except for GMMN + AE which
does not generate images directly but generates hidden layer
activations of an AE that then reconstructs an image. As noted in the
main paper we however want to caution the reader comparing generative
models with respect to log-likelihood as measured by a Parzen-window
estimate can be misleading (see \cite{Theis_2015} for a recent
in-depth discussion).

\begin{table}[h]
  \begin{center}
  \begin{tabular}{lc}
    \bf Algorithm & Log-likelihood\\
    \hline
    GMMN \citep{Li_2015} & 147 $\pm$ 2 \\
    GSN \citep{Bengio_ICML2014} & 214 $\pm$ 1 \\
    GAN \citep{Goodfellow_NIPS2014} & 225 $\pm$ 2 \\
    \textbf{CatGAN} & \textbf{237 $\pm$ 6} \\
    \hline
    GMMN + AE \citep{Li_2015} & \textbf{282 $\pm$ 2}
  \end{tabular}
\end{center}
\caption{Comparison between different generative models on MNIST.}
\label{tab:ll_mnist}
\end{table}

\subsection{Additional plots for experiments on synthetic data}
\begin{figure}[h]
  \includegraphics[width=14cm]{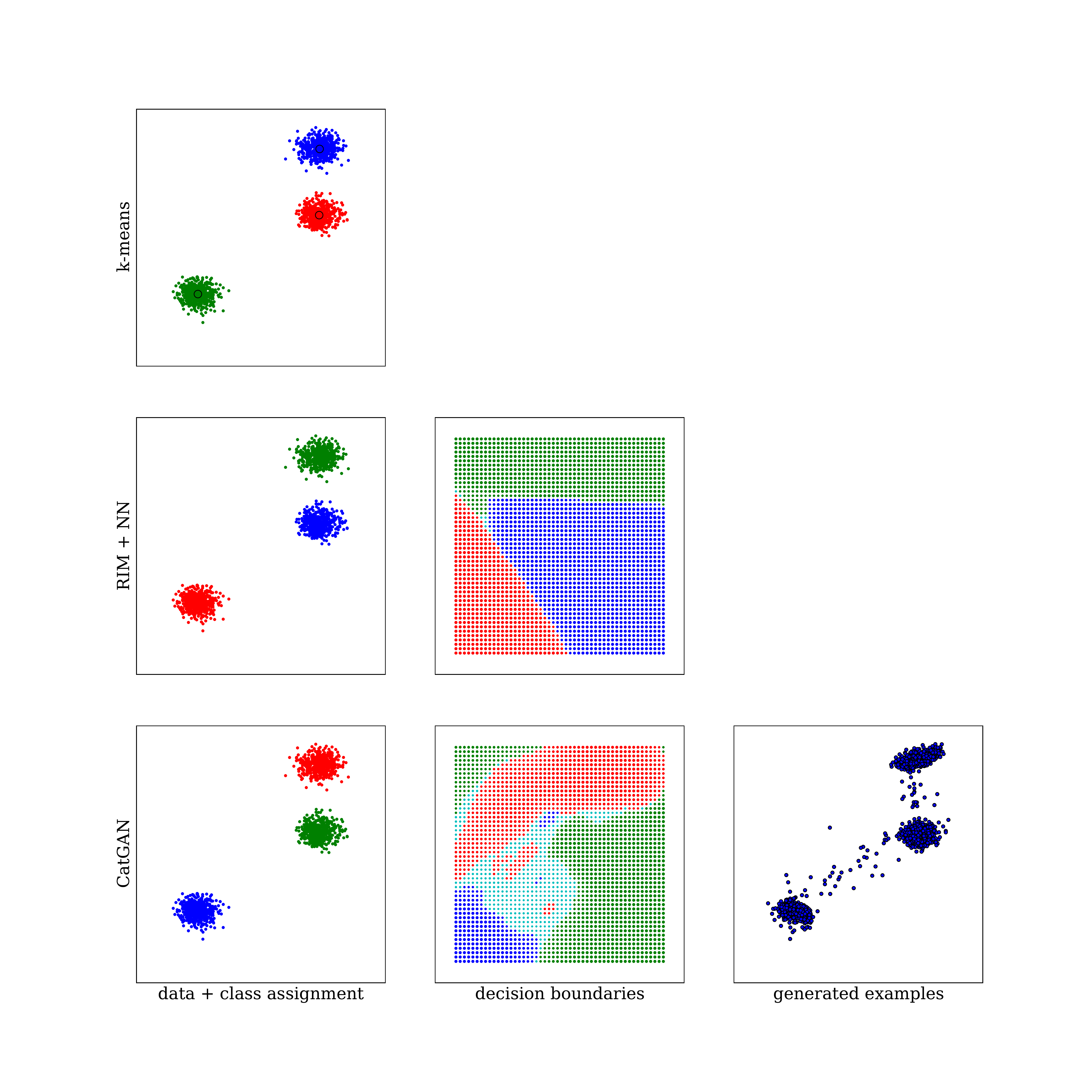}
  \caption{Comparison between k-means, RIM and CatGAN -- with neural
    networks -- on the ``blobs'' dataset, with $K=3$. In the decision boundary
    plots cyan denotes points whose class assignment is close to
    chance level ($\forall k : p(y = k, \bx, D) < 0.55$). Note that the
    class identity is not known a priori as all models are trained
    unsupervisedly (hence the different color/class assignments for
    different models).}
  \label{fig:blobs_compare}
\end{figure}

\begin{figure}[h]
  \includegraphics[width=14cm]{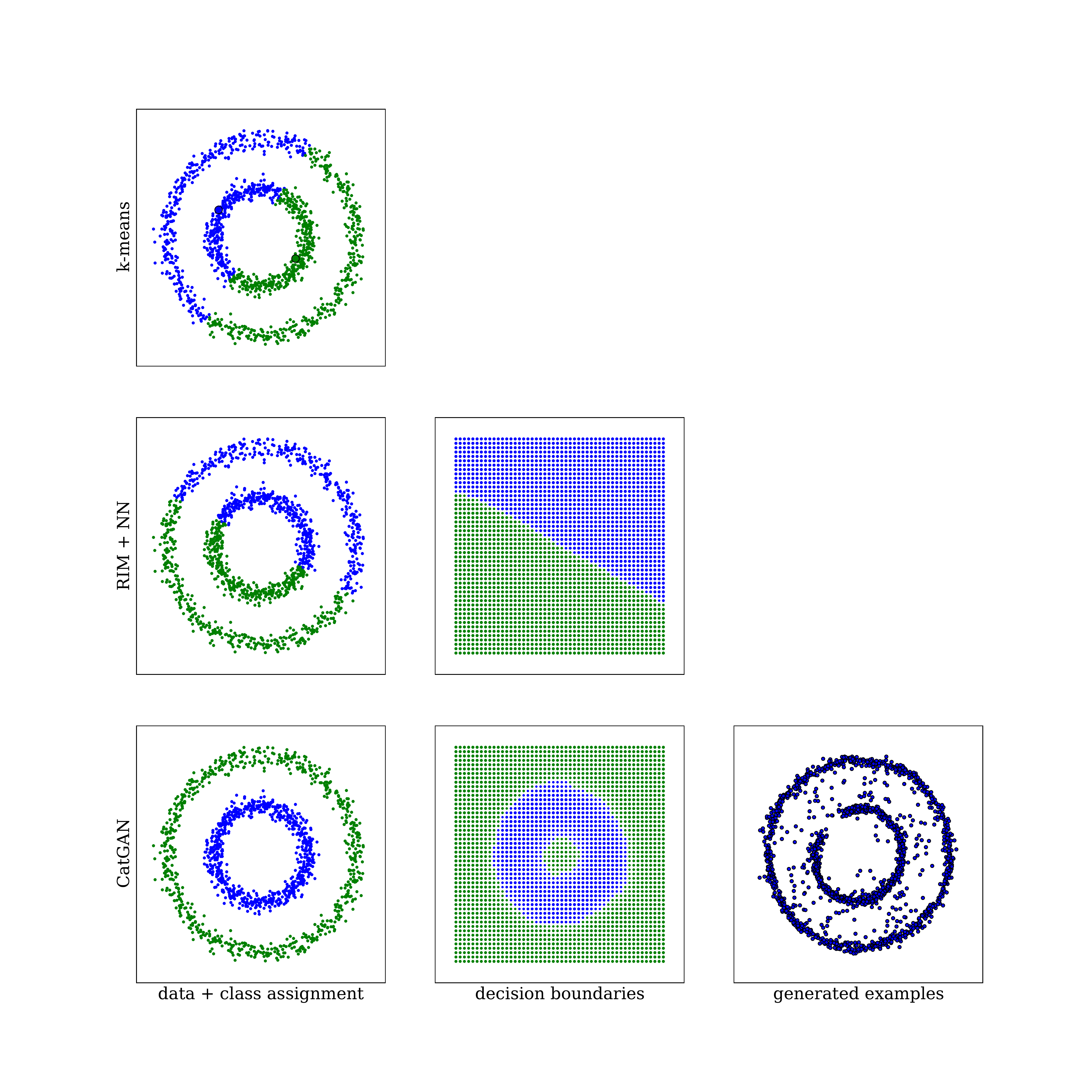}
  \caption{Comparison between k-means, RIM and CatGAN -- with neural
    networks -- on the ``circles'' dataset. This figure complements
    Figure 2 from the main paper.}
  \label{fig:circles_compare}
\end{figure}

In Figure \ref{fig:blobs_compare}, \ref{fig:blobs_compare} and
\ref{fig:circles_compare} we show the results of training k-means, RIM
and CatGAN models on the three synthetic datasets from the main
paper. Only the CatGAN model ``correctly'' clusters the data and, as
an aside, also produces a generative model capable of generating data
points that are almost indistinguishable from those present in the
dataset.  It should be mentioned that there exist clustering
algorithms -- such as DBSCAN \citep{EstKriSanXu96} or spectral clustering
methods -- which can correctly identify the clusters in the datasets
by making additional assumptions on the data distribution.

\begin{figure}[h]
  \includegraphics[width=14cm]{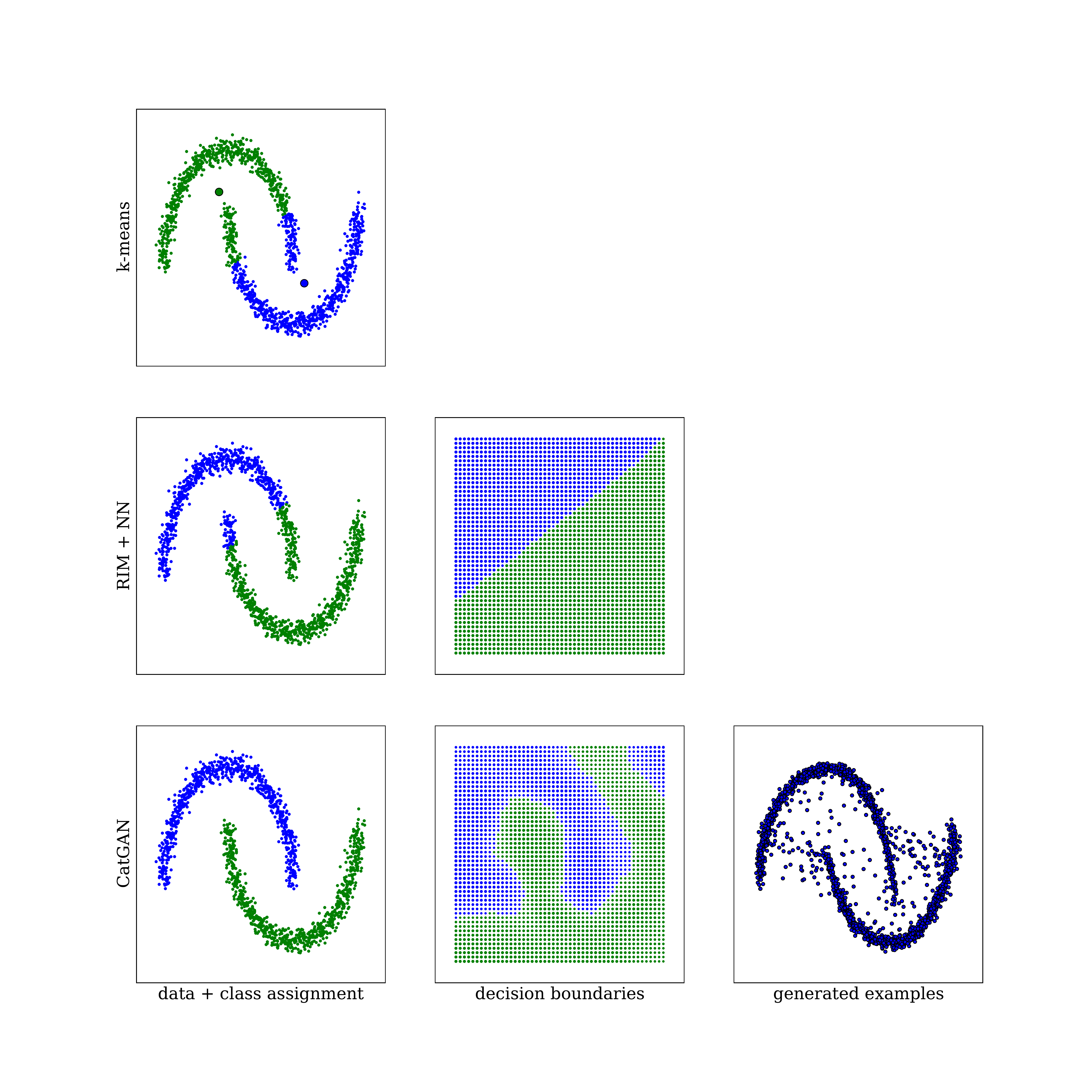}
  \caption{Comparison between k-means, RIM and CatGAN -- with neural
    networks -- on the ``two moons'' dataset, with $K=2$. In the decision boundary
    plots cyan denotes points whose class assignment is close to
    chance level ($\forall k : p(y = k, \bx, D) < 0.55$). Note that the
    class identity is not known a priori as all models are trained
    unsupervisedly (hence the different color/class assignments for
    different models).}
  \label{fig:moons_compare}
\end{figure}

\subsection{Additional visualizations of samples from the generative model}
\begin{figure}[h]
  \centering
  \includegraphics[width=10cm]{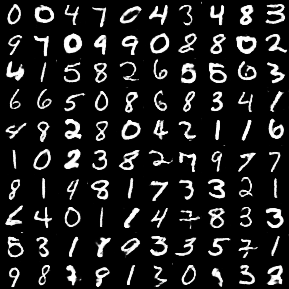}
  \caption{Samples generated by the generator neural network $G$ for a
    CatGAN model trained on the MNIST dataset.}
  \label{fig:mnist_samples_big}
\end{figure}

\begin{figure}[h]
  \includegraphics[width=14cm]{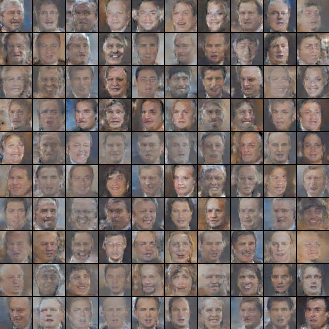}
  \caption{Samples generated by the generator neural network $G$ for a
    CatGAN model trained on cropped images from the LFW dataset.}
  \label{fig:lfw_samples_big}
\end{figure}

We depict additional samples from an unsupervised CatGAN model trained
on MNIST and Labeled Faces in the Wild (LFW)\citep{LFWTech} in Figures
\ref{fig:mnist_samples_big} and \ref{fig:lfw_samples_big}. The
architecture for the MNIST model is the same as in the semi-supervised
experiments and the architecture for LFW is the same as for the
CIFAR-10 experiments.

\end{appendix}

\end{document}